\pdfoutput=1
\PassOptionsToPackage{table}{xcolor}  
\documentclass[11pt]{article}

\usepackage[preprint]{acl}

\usepackage{times}
\usepackage{latexsym}
\usepackage{todonotes}
\usepackage[T1]{fontenc}

\usepackage[utf8]{inputenc}

\usepackage{microtype}

\usepackage{inconsolata}

\usepackage{graphicx}
\usepackage{amsmath}
\usepackage{algorithm}
\usepackage{algorithmic}
\newtheorem{assumption}{Hypothesis}[section]
\usepackage{booktabs}
\usepackage{threeparttable}
\usepackage{multirow}
\usepackage{dcolumn}
\usepackage{amsfonts}
\usepackage{colortbl}

\title{Sparse Brains are Also Adaptive Brains:\\Cognitive-Load-Aware Dynamic Activation for LLMs}

\author{
 \textbf{Yiheng Yang\textsuperscript{1}},
 \textbf{Yujie Wang\textsuperscript{2}}\thanks{wangyujie37@meituan.com},
 \\
 \textbf{Chi Ma\textsuperscript{2}},
 \textbf{Lei Yu\textsuperscript{2}},
 \textbf{Emmanuele Chersoni\textsuperscript{1}},
 \textbf{Chu-Ren Huang\textsuperscript{1}},
\\
 \textsuperscript{1}The Hong Kong Polytechnic University,
 \textsuperscript{2}Meituan,
}

\begin{document}
\maketitle
\begin{abstract}
Dense large language models(LLMs) face critical efficiency bottlenecks as they rigidly activate all parameters regardless of input complexity. While existing sparsity methods(static pruning or dynamic activation) address this partially, they either lack adaptivity to contextual or model structural demands or incur prohibitive computational overhead. Inspired by human brain's dual-process mechanisms - predictive coding (N400) for backbone sparsity and structural reanalysis (P600) for complex context - we propose CLADA, a \textit{\textbf{C}ognitive-\textbf{L}oad-\textbf{A}ware \textbf{D}ynamic \textbf{A}ctivation} framework that synergizes statistical sparsity with semantic adaptability. Our key insight is that LLM activations exhibit two complementary patterns: 1) \textit{Global statistical sparsity} driven by sequence-level prefix information, and 2) \textit{Local semantic adaptability} modulated by cognitive load metrics(e.g., surprisal and entropy). CLADA employs a hierarchical thresholding strategy: a baseline from offline error-controlled optimization ensures 40\%+ sparsity, dynamically adjusted by real-time cognitive signals. Evaluations across six mainstream LLMs and nine benchmarks demonstrate that CLADA achieves \textbf{~20\% average speedup with <2\% accuracy drop}, outperforming Griffin (5\%+ degradation) and TT (negligible speedup). Crucially, we establish the first formal connection between neurolinguistic event-related potential (ERP) components and LLM efficiency mechanisms through multi-level regression analysis ($R^2=0.17$ for sparsity-adaptation synergy). Requiring no retraining or architectural changes, CLADA offers a deployable solution for resource-aware LLM inference while advancing biologically-inspired AI design. Our code is available at \href{https://github.com/Oldify/CLADA}{CLADA}.
\end{abstract}

\section{Introduction \& Related Work}

\begin{figure*}[t]
    \centering
    \includegraphics[width=\linewidth]{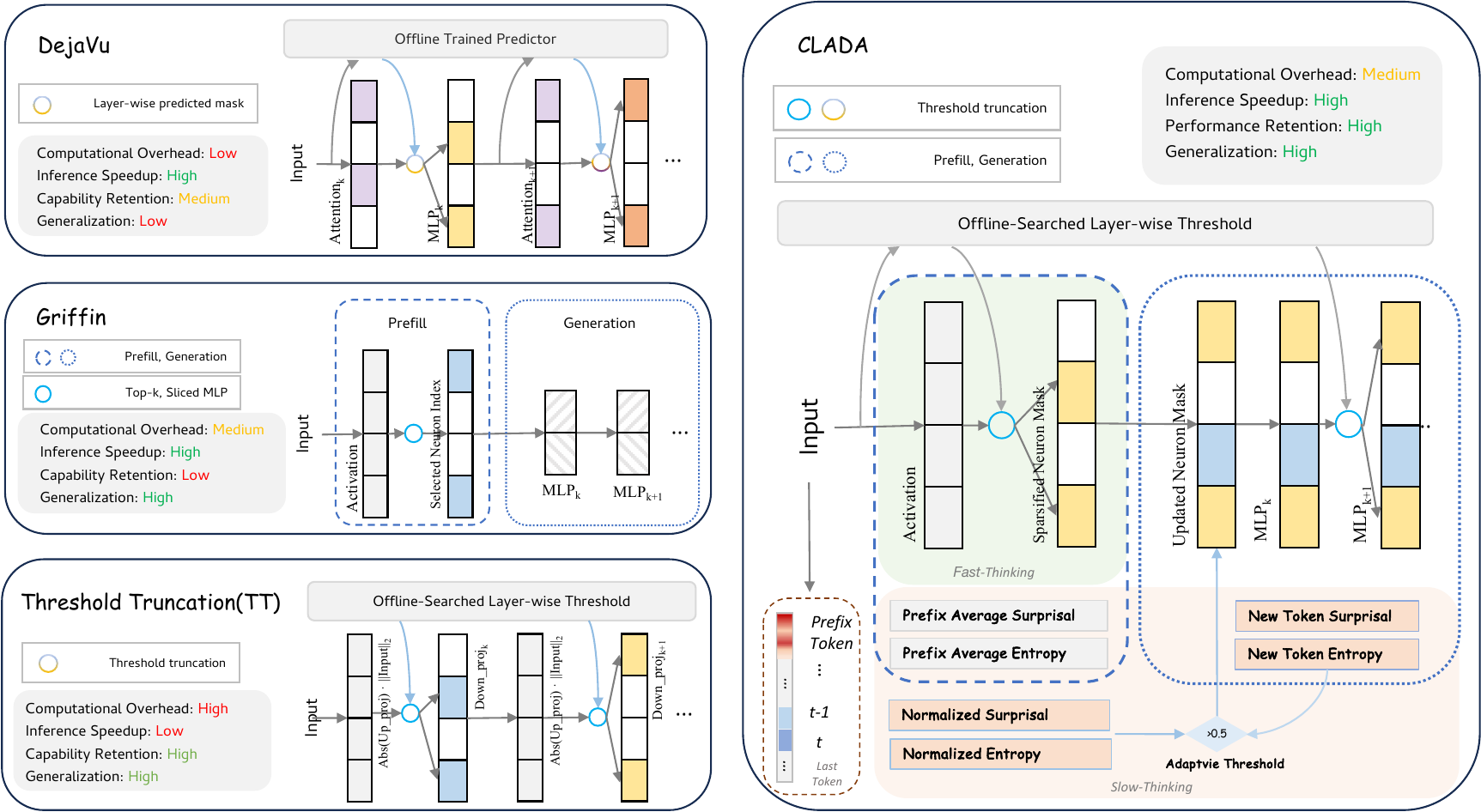}
    \caption{A comparative analysis of dynamic activation methods, highlighting CLADA's advantages in minimal computational overhead, substantial inference speedup, the ability to maximize model capability, and robust generalization ability across various LLM architectures.}
    \label{fig: main}
\end{figure*}

Large Language Models (LLMs) have revolutionized natural language processing through their unprecedented scale and emergent capabilities. Architectures like LLaMA \cite{touvron2023llamaopenefficientfoundation, touvron2023llama2openfoundation, grattafiori2024llama3herdmodels}, Mistral \cite{jiang2023mistral7b}, and OPT \cite{zhang2022optopenpretrainedtransformer} demonstrate remarkable in-context learning abilities that scale with parameter count. However, this performance comes at a steep computational cost: modern LLMs require activating billions of parameters per token during inference, creating critical latency bottlenecks \cite{frantar2023sparsegptmassivelanguagemodels}. While existing sparsity-based optimization methods attempt to address this challenge through static pruning \cite{frantar2023sparsegptmassivelanguagemodels, sun2024simpleeffectivepruningapproach, ashkboos2024slicegptcompresslargelanguage} or mixture-of-experts architectures \cite{zhang2022moeficationtransformerfeedforwardlayers, zhu2024llamamoebuildingmixtureofexpertsllama, szatkowski2024exploitingactivationsparsitydense, zheng2024learnefficientbuildstructured, pan2024densetrainingsparseinference, zhong2024loryfullydifferentiablemixtureofexperts}, they fundamentally lack the dynamic adaptability required for natural language processing. Detailed introduction on more related work can be found at Appendix~\ref{app: related work}.


\begin{table}[t]
\centering
\scalebox{0.68}{
\begin{tabular}{r|cccc}
\toprule
\multicolumn{1}{c}{} & \textbf{MMLU}$\uparrow$ & \textbf{TruthfulQA}$\uparrow$ & \textbf{Winogrande}$\uparrow$ & \textbf{GSM8K}$\uparrow$ \\
\midrule
\multicolumn{1}{c}{\textbf{LLaMA2-7B}} & 45.83 & 61.04 & 74.11 & 13.95 \\
\midrule
DejaVu  & 27.02 & 51.12 & 50.2  & 7.22  \\
TT      & 45.62 & 60.66 & 73.88 & 13.65 \\
Griffin & 43.59 & 59.26 & 73.21 & 12.31 \\
\midrule
\rowcolor{green!5}\textbf{CLADA} & 44.83 & 60.45 & 73.53 & 13.18 \\
\bottomrule
\end{tabular}
}
\caption{Accuracy on benchmarks. Higher values are better. CLADA achieves competitive results on all benchmarks, closely rivaling the baseline LLaMA2-7B.}
\label{table: dejavu}
\end{table}

Furthermore, our analysis reveals three key limitations of current dynamic activation approaches (see Table~\ref{fig: main}): First, the DejaVu\cite{liu2023dejavucontextualsparsity} sparsity framework fails to adapt to no-RELU activation requirements. Second, threshold-based dynamic activation techniques like TT \cite{ma2024dynamicactivationpitfallsllama} incur prohibitive computational overhead that negates their theoretical speedup potential. Third, training-free methods such as Griffin \cite{dong2024promptpromptedadaptivestructuredpruning} exhibit significant performance degradation (see Table~\ref{table: dejavu}) due to their heuristic nature. These shortcomings underscore a fundamental challenge: how to achieve cross-architecture and training-free dynamic computational efficiency while preserving model capacity.

This work draws inspiration from cognitive neuroscience, where the human brain achieves remarkable efficiency through adaptive resource allocation. Neurophysiological studies demonstrate two complementary mechanisms: 1) Predictive coding through the N400 event-related potential (ERP) that minimizes semantic integration costs \cite{li2024decompositionsurprisalunifiedcomputational}, and 2) Structural reanalysis via the P600 ERP that handles syntactic complexity \cite{wilcox2024testingpredictionssurprisaltheory}. Computational analogs of these mechanisms - surprisal \cite{hale-2001-probabilistic, LEVY20081126} and entropy - provide quantitative measures of cognitive load during language processing \cite{salicchi-hsu-2025-every}. The N400-P600 duality in human language processing directly inspires our dual mechanisms: N400-like global sparsity for backbone processing and P600-like local adaptation for complex context.

We posit that LLMs can achieve similar efficiency through two principled sparsity mechanisms:
\begin{enumerate}
    \item \textbf{Statistical Sparsity}: Leveraging sequence-level prefix information to identify and sparsify redundant activations;
    \item \textbf{Semantic Adaptability}: Dynamically allocating computational resources based on real-time cognitive load indicators.
\end{enumerate}

Building on this insight, we propose \textit{\textbf{C}ognitive-\textbf{L}oad-\textbf{A}ware \textbf{D}ynamic \textbf{A}ctivation} (\textbf{CLADA}), a novel framework that synergizes offline error-controlled optimized threshold truncation with cognitive-load-aware adaptation. Our method introduces three key innovations:
\paragraph{Theoretical Foundation}We establish the first connection between cognitive load metrics (surprisal, entropy) and LLM activation patterns, demonstrating their dual role in governing statistical sparsity and semantic adaptation.
\paragraph{Algorithmic Design}CLADA introduces a dynamic thresholding mechanism that automatically adjusts activation masks based on both sequence statistics (prefix-based prediction) and semantic complexity (cognitive-load-aware adaptability), requiring no model retraining.
\paragraph{Empirical Validation}Extensive experiments across six LLM architectures and nine benchmarks reveal that CLADA achieves 20\% average speedup with <2\% performance degradation (Table~\ref{table: dejavu} and~\ref{table: latency}), while outperforming existing methods in computational overhead and generalization (Table~\ref{fig: main}).

Our work bridges cognitive science with machine learning, offering a new paradigm for designing efficient yet interpretable language models. By grounding dynamic activation strategies in cognitive theory, we take a significant step toward building LLMs that mirror the adaptive efficiency of biological neural systems.

\section{Bridging Cognitive Load and Activation}

\subsection{Global Statistical Sparsity}
\label{sec: statistical sparsity}
\paragraph{Theoretical Framework}
Statistical sparsity characterizes the phenomenon where neuron activation patterns exhibit hierarchical dependence on sequence prefixes. We formalize this through Hypothesis~\ref{hypo1}:

\begin{assumption}
\label{hypo1}
    (\textbf{Existence}) Statistical sparsity manifests as prefix-dependent activation flocking, where activation matrix similarity between hybrid sequences grows monotonically with consistent prefix length.
\end{assumption}

This hypothesis aligns with hierarchical representation theories in language modeling \cite{dubey2022activationfunctionsdeeplearning, guo2024attentionscoreneedtoken}, where early tokens establish persistent activation patterns. To operationalize this, we develop a panel regression framework:

\begin{equation}
    \Delta_{\text{sim}} = \mu_i + \beta\ell_{ij} + \alpha_{i} + \epsilon_{ij}
    \label{eq: reg1}
\end{equation}
where:
\begin{itemize}
    \item $\mu_i$: Constant term. 
    \item $\ell_{ij}$: Prefix length, the subscript $j$ represents data with varying prefix ratios
    \item $\alpha_i$: The individual fixed effect term in panel data
    \item $\epsilon_{ij}$: Random perturbation error term
\end{itemize}

\subsubsection{Experimental Design}
Our multi-stage validation framework employs controlled sequence generation and activation pattern analysis:

\paragraph{Corpus Construction}
Using XSum \cite{narayan2018dontdetailsjustsummary}, we create:
\begin{itemize}
    \item \textbf{N}atural \textbf{L}anguage \textbf{S}equences (\textbf{NLS}): 2,000 random authentic text samples
    \item \textbf{R}andom \textbf{T}oken \textbf{S}equences (\textbf{RTS}): 2,000 sequences generated by permuting XSum vocabularies while preserving part-of-speech ratios and punctuation frequency
\end{itemize}

\paragraph{Hybrid Sequence Generation}\label{subsec: hybrid}
For each prefix ratio $\alpha \in \{0.25, 0.30, 0.35, 0.40, 0.45, 0.50\}$:
\begin{enumerate}
    \item Compute prefix length: $\ell = \lceil2048\alpha\rceil$ ($512 \leq \ell \leq 1024$).
    \item Create hybrid sequences A' by replacing first $\ell$ tokens of NLS-A/RTS-B with NLS-B/RTS-B counterparts.
    \item Generate six experimental groups (36,000 sequences in total):
    \begin{itemize}
        \item NLS-A, NLS-B, NLS-A'
        \item RTS-A, RTS-B, RTS-A'
    \end{itemize}
\end{enumerate}

\subsubsection{Activation Matrix Extraction}
For each sequence, we extract Layer-15 activations from LLaMA-3-8B \cite{grattafiori2024llama3herdmodels} at Token-2049:
\[
\mathbf{M}^{(l=15)} \in \mathbb{R}^{1 \times (4096 \times 14336)}
\]
The layer selection aligns with previous findings on mid-layer semantic integration \cite{skean2025layerlayeruncoveringhidden}. The rationale for selecting token-2049 is detailed in Appendix \ref{app: token}.

\subsubsection{Similarity Metric Calculation}
We compute normalized activation similarity shift:
\begin{equation}\label{eq: sim}
    \Delta_{\text{sim}} = \frac{\text{sim}(\mathbf{M}_{A'}, \mathbf{M}_B)}{\text{sim}(\mathbf{M}_A, \mathbf{M}_B)} - 1
\end{equation}
where similarity is measured via centered kernel alignment (CKA) \cite{cortes2024algorithmslearningkernelsbased}:
\begin{equation}\label{eq: sim_cka}
    \text{sim}_{\text{CKA}}(\mathbf{X},\mathbf{Y}) = \frac{\|\mathbf{X}^\top\mathbf{Y}\|_F^2}{\|\mathbf{X}^\top\mathbf{X}\|_F \|\mathbf{Y}^\top\mathbf{Y}\|_F}
\end{equation}
To ensure the robustness of the regression, we additionally employed cosine similarity to quantify the similarity between activation matrices:
\begin{equation}\label{eq: sim_cos}
    \text{sim}_{\text{cosine}}(\mathbf{X}, \mathbf{Y}) = \frac{\mathbf{X}^\top \mathbf{Y}}{\|\mathbf{X}\|_2 \|\mathbf{Y}\|_2}
\end{equation}

\begin{algorithm}[t]
\caption{Empirical Validation of the Existence of Statistical Sparsity}
\label{alg: existence}
\begin{algorithmic}[1]
\REQUIRE Sequences $A$, $B$; Prefix ratio $\alpha$
\STATE $A' \gets \text{generate\_hybrid\_sequence}(A, B, \alpha)$
\STATE $\mathbf{M}_A \gets \text{extract\_activations}(A)$
\STATE $\mathbf{M}_B \gets \text{extract\_activations}(B)$
\STATE $\mathbf{M}_{A'} \gets \text{extract\_activations}(A')$

\STATE $\text{sim}_{AB} \gets \text{sim}(\mathbf{M}_A, \mathbf{M}_B)$
\STATE $\text{sim}_{A'B} \gets \text{sim}(\mathbf{M}_{A'}, \mathbf{M}_B)$
\RETURN $\Delta_{\text{sim}} = \frac{\text{sim}(\mathbf{M}_{A'}, \mathbf{M}_B)}{\text{sim}(\mathbf{M}_A, \mathbf{M}_B)} - 1$
\end{algorithmic}
\end{algorithm}

\subsubsection{Empirical Validation}
\paragraph{Experiment 1.1: RTS Groups}
Following Algorithm~\ref{alg: existence}, construct hybrid sequence $A'$ from RTS $B$ according to Section~\ref{subsec: hybrid}. Activation matrix similarity is computed using Equation \ref{eq: sim}.

\begin{table}[htbp]
    \centering
    \scalebox{0.58}{
    \begin{threeparttable}
    \begin{tabular}{lcccccc}
    \toprule
    ~ & \multicolumn{6}{c}{Dependent variable: $\Delta_{\text{CKA\_sim}}$} \\
    \cmidrule(lr){2-7}
    ~ & (1) & (2) & (3) & (4) & (5) & (6) \\
    ~ & RTS-A' & NLS-A' & RTS-A' & NLS-A' & RTS-A' & NLS-A' \\
    \midrule
    Prefix\_len & $4.12^{***}$ & $3.91^{***}$ & $4.05^{***}$ & $4.03^{***}$ & $3.75^{***}$ & $3.69^{***}$ \\
              ~ & (22.03) & (20.16) & (21.53) & (21.37) & (19.22) & (18.84) \\
    Surprisal   & ~ & ~ & -0.02 & $-0.85^{***}$ & -0.09 & $-0.80^{***}$ \\
              ~ & ~ & ~ & (-0.26) & (-8.19) & (-0.14) & (-8.13) \\
    Entropy & ~ & ~ & ~ & ~ & -0.36 & $-0.12^{***}$ \\
              ~ & ~ & ~ & ~ & ~ &(-4.38) &(-5.87) \\
    \midrule
    Token\_len & $10.38^{***}$ & $13.90^{***}$ & $10.14^{***}$ & $12.31^{***}$ & $11.23^{***}$ & $11.38^{***}$ \\
             ~ & (5.06) & (5.06) & (5.07) & (5.05) & (5.07) & (5.06)  \\
    \midrule
    Obs & 12000 & 12000 & 12000 & 12000 & 12000 & 12000 \\
    Adjusted R$^2$ & 0.13 & 0.12 & 0.15 & 0.16 & 0.17 & 0.16 \\
    \midrule
    Individual FE & YES & YES & YES & YES & YES & YES \\
    Constant & YES & YES & YES & YES & YES & YES \\
    \bottomrule
    \end{tabular}
    \begin{tablenotes}
    \small
    \item Note: * p<0.1; ** p<0.05; *** p<0.01
    \end{tablenotes}
    \end{threeparttable}
    }
    \caption{Regression results on the impact of \textit{prefix\_len} on CKA activation similarity}
    \label{table: reg}
\end{table}

\textbf{Observation 1}: Regression results (Table~\ref{table: reg} Column 1) reveal strong prefix-length effects ($\beta=4.12, p<0.001$), confirming Hypothesis~\ref{hypo1}. The robustness check with cosine similarity (Table~\ref{table: robust}) yields consistent outcomes ($\beta=4.22, p<0.001$).

\paragraph{Experiment 1.2: NLS Groups}
Replicate the aforementioned experimental procedures in Experiment 1.1, regression results of NLS groups are reported in the second column of Table~\ref{table: reg}.

\begin{table}[htbp]
    \centering
    \scalebox{0.58}{
    \begin{threeparttable}
    \begin{tabular}{lcccccc}
    \toprule
    ~ & \multicolumn{6}{c}{Dependent variable: $\Delta_{\text{cos\_sim}}$} \\
    \cmidrule(lr){2-7}
    ~ & (1) & (2) & (3) & (4) & (5) & (6) \\
    ~ & RTS-A' & NLS-A' & RTS-A' & NLS-A' & RTS-A' & NLS-A' \\
    \midrule
    Prefix\_len & $4.22^{***}$ & $3.94^{***}$ & $4.08^{***}$ & $4.03^{***}$ & $3.71^{***}$ & $3.64^{***}$ \\
              ~ & (22.12) & (20.41) & (21.35) & (21.15) & (19.46) & (18.53) \\
    Surprisal   & ~ & ~ & -0.01 & $-0.72^{***}$ & -0.07 & $-0.80^{***}$ \\
              ~ & ~ & ~ & (-0.26) & (-7.32) & (-0.12) & (-8.14) \\
    Entropy & ~ & ~ & ~ & ~ & -0.35 & $-0.12^{***}$ \\
              ~ & ~ & ~ & ~ & ~ &(-4.25) &(-5.85) \\
    \midrule
    Token\_len & $10.63^{***}$ & $13.35^{***}$ & $10.65^{***}$ & $12.64^{***}$ & $11.75^{***}$ & $11.34^{***}$ \\
             ~ & (5.04) & (5.07) & (5.02) & (5.05) & (5.06) & (5.03)  \\
    \midrule
    Obs & 12000 & 12000 & 12000 & 12000 & 12000 & 12000 \\
    Adjusted R$^2$ & 0.12 & 0.12 & 0.15 & 0.16 & 0.17 & 0.16 \\
    \midrule
    Individual FE & YES & YES & YES & YES & YES & YES \\
    Constant & YES & YES & YES & YES & YES & YES \\
    \bottomrule
    \end{tabular}
    \begin{tablenotes}
    \small
    \item Note: * p<0.1; ** p<0.05; *** p<0.01
    \end{tablenotes}
    \end{threeparttable}
    }
    \caption{Robustness Test: Regression results of \textit{prefix\_len} on cosine activation similarity}
    \label{table: robust}
\end{table}

\textbf{Observation 2}: While maintaining statistical significance ($\beta=3.91,p<0.001$) (Table~\ref{table: reg} Column 2), the attenuated effect size compared to RTS ($\Delta\beta=5.4\%$) suggests cognitive-load-aware interference in natural language processing. This observation motivates our investigation of semantic adaptability.

\subsubsection{Visual Evidence and Case Study}
To investigate the existence of Statistical Sparsity in a visual and intuitive way, we extracted randomly two entry from XSum dataset as \textit{\textbf{N}atural \textbf{L}anguage \textbf{S}equence-\textbf{A}} (\textbf{NLS-A}) and NLS-B, and replaced the first 512 tokens of NLS-A with NLS-B to produce NLS-A'. Details can be seen in Appendix \ref{app: visual}. And a sentence-by-sentence case study can be seen in Appendix \ref{app: detailed samples}

\subsection{Local Semantic Adaptability: Cognitive-Load-Aware Response}
\label{sec: semantic adapt}
\paragraph{Theoretical Foundation}
The NLS-RTS performance gap ($4.12$ vs $3.91$) reveals a critical limitation of pure statistical sparsity: cognitive load disrupts prefix-dependent activation patterns. We formalize this complementary mechanism through Hypothesis~\ref{hypo2}:

\begin{assumption}
\label{hypo2}
    (\textbf{Existence}) Semantic adaptability emerges through cognitive-load-modulated activation adjustments, quantified by surprisal($s_t$) and entropy($H_t$).
\end{assumption}
 
\subsubsection{Operationalization}
We extend the regression framework to incorporate cognitive load metrics:
\begin{equation}
    \Delta_{\text{sim}} = \mu_i + \beta\ell_{ij} + \gamma_1 s_{ijt} + \gamma_2 H_{ijt} +\alpha_{i} + \epsilon_{ij}
    \label{eq: reg2}
\end{equation}

\paragraph{Cognitive Load Quantification}
Following psycholinguistic theory \citep{hale2016information,oh2024frequencyexplainsinversecorrelation,salicchi-hsu-2025-every}:
\begin{align}
    \label{align: surprisal} s_t &= -\log P(w_t | w_{<t}) \quad \text{(surprisal)} \\
    \label{align: entropy} H_t &= -\sum_{w \in \mathcal{V}} P(w|w_{<t}) \log P(w|w_{<t}) \quad \text{(entropy)}
\end{align}
where:
\begin{itemize}
    \item $w_t$: the target token
    \item $w_{<t}$: the context of target token
    \item $P(w_t | w_{<t})$: the conditional probability of target word $w_t$ provided by the language model
    \item $\mathcal{V}$: the vocabulary of the language model enriched with special token (e.g., EOS)
\end{itemize}
And the descriptive statistics of our dataset are reported in Table~\ref{table: descriptive} in Appendix~\ref{sec: details}.

\subsubsection{Empirical Validation}
\paragraph{Experiment 2: Cognitive-Load-Aware Interference}
Key findings from Table~\ref{table: reg} Columns 3-6:
\begin{itemize}
    \item \textbf{RTS}: Non-significant cognitive-load-aware effects($\gamma_1=-0.07, \gamma_2=-0.35,p>0.1$)
    \item \textbf{NLS}: Significant inverse correlations($\gamma_1=-0.80, \gamma_2=-0.12, p<0.001$)
\end{itemize}

This demonstrates semantic adaptability's domain-specific nature: high surprisal/entropy tokens in NLS activate emergent neurons (ENs), disrupting prefix-driven activation patterns. The robustness analysis (Table~\ref{table: robust} Column 3-6) confirms these effects persist across similarity metrics.

\section{Methodology}
\label{sec:methodology}

\subsection{Unified Framework: cognitive-Load-Aware Mechanism}
The interplay between statistical sparsity and semantic adaptability forms a \textit{hierarchical activation mechanism}, where global efficiency is balanced by local precision. 

\paragraph{Core Design} 
\begin{equation}
  A_{\text{total}}^{(t)} = \underbrace{A_{\text{base}}^{(t)}}_{\text{Statistical Sparsity}} + \underbrace{\Delta A_{\text{semantic}}^{(t)}}_{\text{Semantic Adaptability}}
\end{equation}
where:
\begin{itemize}
    \item Statistical Sparsity (\(A_{\text{base}}^{(t)}\)): Provides a stable, sequence-level efficiency baseline.
    \item Semantic Adaptability (\(\Delta A_{\text{semantic}}^{(t)}\)): Dynamically adjusts activation intensity to handle high-cognitive-load contexts.
\end{itemize}



This dual mechanism aligns with the human brain’s dual-system processing: \textit{System 1} (fast, pattern-driven) and \textit{System 2} (slow, context-aware) \cite{saha2024system1xlearningbalancefast}.

\subsection{Dynamic Threshold Strategy}
\subsubsection{Neuron Activation Prefilling}
For each layer $l$, compute neuron-wise activation magnitudes during \textit{prefilling} phase:
\begin{equation}
\begin{split}
    A_j^{(l)} =& \left\| \left[ \sigma(W_{\text{in},j}^{(l)}x) \odot V_{\text{in},j}^{(l)}x \right] W_{\text{out},j}^{(l)} \right\|_2 \\ & \quad \forall j \in [1, d_h^{(l)}]
\end{split}
\end{equation}
\label{eq: neuron_activation}
where $j$ indexes neurons in layer $l$'s hidden dimension $d_h^{(l)}$. $W_{\text{in},j}^{(l)}$ and $V_{\text{in},j}^{(l)}$ denote the $j$-th row of input projection matrices. $W_{\text{out},j}^{(l)}$ is the $j$-th column of output projection matrix.

\subsubsection{Layer-wise Base Threshold}
Determine optimal base thresholds through error-controlled optimization searching:
\begin{equation}
\begin{aligned}
    \tau_{\text{base}}^{(l)} &= \arg\max_{\epsilon^{(l)}} \epsilon^{(l)} \\
    \text{s.t.}\quad \mathbb{E}_{\mathbf{x} \sim \mathcal{D}_{\text{val}}} & \left[ \text{CETT}^{(l)}(\mathbf{x}; \epsilon^{(l)}) \right] \leq 0.2 \\
    \text{where}\quad \text{CETT}^{(l)} &= \frac{\left\| \sum_{j \in \mathcal{S}_{\text{cut}}^{(l)}} n_j^{(l)}(\mathbf{x}) \right\|_2}{\left\| MLP^{(l)}(\mathbf{x}) \right\|_2} \\
    \mathcal{S}_{\text{cut}}^{(l)} &= \left\{ j \;\big|\; A_j^{(l)} < \epsilon^{(l)} \right\}
\end{aligned}
\label{eq: base_threshold}
\end{equation}

\subsubsection{Token-aware Threshold Adjustment}
Integrate token-level cognitive signals with layer-wise base thresholds in Equation~\ref{eq: final_threshold}
\begin{figure*}[h]
\begin{equation}
    \tau_{\text{final}}^{(l)}(t) = \tau_{\text{base}}^{(l)} \cdot \left[ 1 + \lambda^{(l)} \cdot \mathbb{I}(s_t > \tau_s) + \gamma^{(l)} \cdot \mathbb{I}(H_t > \tau_H) \right]
    \label{eq: final_threshold}
\end{equation}
\end{figure*}
where token-level metrics are computed as Equation~\ref{align: surprisal} and Equation~\ref{align: entropy}.

\begin{itemize}
    \item $\mathbb{I}(\cdot)$: Indicator function (1 if condition met, 0 otherwise)
    \item $\tau_s, \tau_H$: Universal thresholds across all layers
    \item $\lambda^{(l)}, \gamma^{(l)}$: Layer-specific scaling coefficients. Here, we choose 0.80 and 0.12 from Table~\ref{table: reg}
\end{itemize}
Algorithm Implementation can be found at Algorithm~\ref{alg:CLADA}.

\begin{algorithm}[t]
\caption{CLADA Inference}
\label{alg:CLADA}
\textbf{Input}: Model $\theta$, prefix $x_{1:T}$, max gen\_len $N$. \\
\textbf{Output}: Generated sequence $y_{1:N}$. \\
\begin{algorithmic}[1]
\STATE \textbf{Prefill Phase}:
\STATE Search $\tau_{\text{base}}^{(l)}$ and fit $\lambda^{(l)}, \gamma^{(l)}$ for each layer $l$.
\STATE Generate initial activation masks $\{\text{Mask}^{(l)}\}$ using $\tau_{\text{base}}^{(l)}$.

\STATE \textbf{Generation Phase}:
\FOR {$t = 1$ to $N$}
    \STATE Compute $s(w_{<T}^{t-1})$ and $H(w_{<T}^{t-1})$.
    \STATE Update thresholds:$\tau_{\text{final}}^{(l)}(t)$
    \STATE Regenerate masks $\{\text{Mask}^{(l)}\}$ by thresholding $A_l$.
    \STATE Perform sparse forward pass using $\{\text{Mask}^{(l)}\}$ to predict $y_t$.
    \STATE Update context: $w_{<T}^t = w_{<T}^{t-1} \cup \{y_t\}$.
\ENDFOR
\end{algorithmic}
\end{algorithm}

\section{Experiments}\label{section:experiments}
\subsection{Experimental Setup}
Our approach, along with the baseline models, is implemented using the PyTorch framework, and we leverage the Hugging Face Transformers library for model and dataset management. Our experiments are powered by 1 NVIDIA A100 GPUs with 80 GB of memory. Adhering to the methodologies outlined in Section \ref{sec:methodology}, we sequentially applied our methods for each Transformer layers, which reduces inference latency while preserving model performance. All experiments are conducted in a single phase, without any post-training or fine-tuning stages.

\paragraph{Models} 
In this paper, we conducted a comprehensive series of experiments using the OPT-350M, OPT-2.7B, Gemma-2B, LLaMA-2-7B and LLaMA-3-8B and Mistral-7B models. These models represent a significant advancement in language modeling capabilities, providing a spectrum of scales to meet various computational needs and performance benchmarks.

\paragraph{Tasks and Datasets}
Following Griffin \cite{dong2024promptpromptedadaptivestructuredpruning}, we conduct evaluations on a variety of models across multiple generation and classification tasks. For generation tasks, we focus on XSum\cite{narayan2018dontdetailsjustsummary}, CNN/DailyMail\cite{nallapati2016abstractivetextsummarizationusing}, COQA\cite{reddy2019coqaconversationalquestionanswering}, and QASPER\cite{shaham2022scrollsstandardizedcomparisonlong}. For classification tasks, our evaluation includes HellaSwag\cite{zellers2019hellaswagmachinereallyfinish}, PIQA\cite{bisk2019piqareasoningphysicalcommonsense}, COPA\cite{roemmele2011choice}, ARC-Challenge\cite{clark2018thinksolvedquestionanswering}, and BoolQ\cite{clark2019boolqexploringsurprisingdifficulty}. Except for XSum and CNN/DailyMail, our experiments utilize the LM Evaluation Harness\cite{eval-harness}. 

\paragraph{Baselines}
Besides comparing against the original LLM, we also evaluate CLADA in relation to Griffin and TT. Unless specified otherwise, each technique is applied in a layer-wise manner, enhancing scalability even when dealing with exceptionally large models. TT has similary performance with CLADA, therefore we only evaluate its generation phase latency. For DejaVu, we did not consider it as a comparable baseline in subsequent experiments(see Table~\ref{table: dejavu}).

\paragraph{Sparsity}
In our evaluation, we especially focus on the MLP blocks of LLM models, which constitute approximately 67\% of the parameters of model's two main blocks, making them a crucial target for dynamic activation. Griffin and CLADA achieves around 50\% of sparsity in total.

\subsection{Performance Evaluation}
Table \ref{table: performance} delineates the performance differences between the Griffin and CLADA methods across various generation and classification tasks. Metrics such as Accuracy (Acc), Rouge-1, and F1 scores were measured across various datasets.

\begin{table*}[]
\centering
\scalebox{0.95}{
\begin{tabular}{c|ccccc|cc|cc}
\toprule
 &
  \multicolumn{5}{c}{\textbf{Acc}} &
  \multicolumn{2}{c}{\textbf{Rouge-1}} &
  \multicolumn{2}{c}{\textbf{F1}} \\
  \midrule
\textbf{Models} &
  \textbf{Hellaswag} &
  \textbf{Piqa} &
  \textbf{Copa} &
  \textbf{Arc-c} &
  \textbf{Boolq} &
  \textbf{Xsum} &
  \textbf{Cnn} &
  \textbf{Coqa} &
  \textbf{Qasper} \\
  \midrule
\rowcolor{orange!10}\textbf{OPT-350M}   & 32.06 & 64.64 & 72.00 & 21.33 & 41.01 & 12.89 & 14.82 & 33.39 & 3.34  \\
Griffin    & 30.52 & 62.46 & 69.00 & 20.24 & 39.71 & 10.59 & 13.32 & 31.89 & 2.14  \\
\rowcolor{green!5}CLADA        & 32.00 & 64.04 & 72.00 & 20.73 & 40.76 & 11.23 & 13.47 & 32.24 & 2.45  \\
\midrule
\rowcolor{orange!10}\textbf{OPT-2.7B}   & 45.86 & 73.78 & 77.00 & 60.77 & 66.79 & 18.43 & 22.24 & 64.41 & 7.85  \\
Griffin    & 43.76 & 71.84 & 76.00 & 58.21 & 65.92 & 17.43 & 20.74 & 62.91 & 6.85  \\
\rowcolor{green!5}CLADA        & 45.74 & 73.18 & 76.00 & 58.42 & 66.19 & 17.86 & 21.33 & 64.05 & 7.70  \\
\midrule
\rowcolor{orange!10}\textbf{Gemma-2B}   & 71.40 & 77.30 & 83.00 & 42.10 & 69.40 & 15.69 & 23.32 & 72.03 & 12.46 \\
Griffin    & 70.03 & 76.34 & 82.00 & 41.19 & 68.42 & 14.69 & 22.18 & 71.78 & 11.83 \\
\rowcolor{green!5}CLADA        & 70.85 & 76.21 & 82.00 & 41.19 & 68.21 & 15.32 & 22.51 & 72.45 & 12.33 \\
\midrule
\rowcolor{orange!10}\textbf{LLaMA-2-7B} & 57.16 & 78.07 & 87.00 & 43.34 & 77.71 & 27.15 & 10.08 & 77.35 & 26.31 \\
Griffin    & 56.66 & 76.57 & 85.00 & 41.84 & 76.21 & 26.65 & 8.58  & 75.85 & 25.81 \\
\rowcolor{green!5}CLADA        & 56.86 & 77.67 & 86.00 & 42.84 & 77.51 & 26.85 & 9.98  & 76.95 & 26.11 \\
\midrule
\rowcolor{orange!10}\textbf{LLaMA-3-8B} & 62.53 & 81.85 & 93.00 & 46.29 & 80.76 & 29.62 & 12.21 & 82.92 & 28.86 \\
Griffin    & 62.03 & 80.35 & 91.00 & 43.79 & 78.26 & 27.12 & 11.71 & 82.42 & 27.36 \\
\rowcolor{green!5}CLADA        & 62.31 & 81.40 & 92.00 & 45.79 & 80.39 & 29.47 & 11.93 & 82.57 & 28.37 \\
\midrule
\rowcolor{orange!10}\textbf{Mistral-7B} & 61.21 & 80.58 & 92.00 & 50.43 & 83.61 & 28.67 & 28.00 & 80.70 & 24.56 \\
Griffin    & 59.71 & 79.08 & 92.00 & 47.43 & 82.11 & 27.17 & 26.50 & 78.20 & 22.06 \\
\rowcolor{green!5}CLADA        & 59.32 & 79.21 & 92.00 & 49.24 & 83.14 & 28.35 & 27.53 & 80.55 & 24.07 \\
\bottomrule
\end{tabular}
}
\caption{Generation and classification performance across various model architectures. Higher values are better.}
\label{table: performance}
\end{table*}

Our comprehensive evaluation across six model architectures and nine benchmarks reveals three key findings regarding CLADA's effectiveness:
\paragraph{Consistent Superiority Across Scales.}CLADA outperforms Griffin across all model sizes:
\begin{itemize}
    \item Small(OPT-350M): $+1.48\%$ on Hellaswag
    \item Medium(OPT-2.7B): $+1.34\%$ on Piqa
    \item Larger(LLaMA-3-8B): $+1.30\%$ on BoolQ
\end{itemize}

\paragraph{Task-Agnostic Benefits.}The advantages manifest across both discriminative and generative tasks:
\begin{itemize}
    \item Classification: up to $2\%$ higher accuracy on reasoning-heavy benchmarks(ARC-C, Coqa)
    \item Generation: up to $2.35\%$ better Rouge-1/F1 on summarization(XSum, CNN)
\end{itemize}

\paragraph{Scaling Characteristics.}The performance gap widens with model capacity. Griffin employs a hard top-k, which may discard some important neurons. In contrast, CLADA uses a cognitive-load-aware thresholds, providing greater adaptability.





\subsection{Efficiency Evaluation}
Table \ref{table: latency} provides a comparative analysis of the generation latency for various models on a single NVIDIA A100 GPU, using a batch size of 1 and models implemented in FP16 precision via Hugging Face. 
Both the prompt length and the generated new token length are set to $1024$(longer context results reported in Table~\ref{table:4k}, and the sparsity of Griffin and CLADA both set to 50\%. The unit of reported numbers in Table \ref{table: latency} is seconds.

\begin{table*}[h]
\centering
\scalebox{1.0}{
\begin{tabular}{c|ccc|c}
\toprule
\textbf{Models} & \textbf{Dense} & \textbf{TT} & \textbf{Griffin} & \textbf{CLADA}   \\
\midrule
OPT-2.7B    & 32.95 & 33.52 & 26.96(22.22\%$\downarrow$) & 27.77(18.65\%$\downarrow$)  \\
Gemma-2B    & 30.17 & 30.16 & 23.92(26.13\%$\downarrow$) & 24.06(25.39\%$\downarrow$)  \\
LLaMA-3-8B  & 81.31 & 79.88 & 63.32(22.13\%$\downarrow$) & 64.03(21.25\%$\downarrow$)  \\
Mistral-7B  & 79.28 & 76.26 & 63.26(25.32\%$\downarrow$) & 63.94(19.34\%$\downarrow$)  \\
\bottomrule
\end{tabular}
}
\caption{Generation phase latency(s).}
\label{table: latency}
\end{table*}

The results demonstrate that the CLADA method consistently reduces generation latency compared to the dense configuration across all evaluated models. As shown in Table~\ref{table: latency}, both Griffin and CLADA offer great speedups, ranging from 18-25\%, whereas TT maintains a similar generation latency to dense models.

Overall, above results underscore the efficiency of CLADA in accelerating generation speed without significantly compromising task performance, making it a practical and effective solution for optimizing LLMs.

\begin{table*}[htbp]
\centering
\scalebox{0.85}{
\begin{tabular}{l|cccc|cc}
\toprule
\textbf{Variant} & \multicolumn{4}{c|}{\textbf{Accuracy Retention (\%)}} & \multicolumn{2}{c}{\textbf{Latency (s)}} \\
\cmidrule(lr){2-5} \cmidrule(lr){6-7}
 & \textbf{HellaSwag} & \textbf{PIQA} & \textbf{ARC-C} & \textbf{XSum} & \textbf{OPT-2.7B} & \textbf{LLaMA-3-8B} \\
\midrule
\textbf{CLADA (Full)} & 99.6 & 99.5 & 98.9 & 99.5 & 27.8 & 64.0 \\
w/o Stat. Sparsity & 80.4 & 82.1 & 78.3 & 76.3 & 26.9 & 63.3 \\
w/o Sem. Adapt. & 97.1 & 97.7 & 96.5 & 98.3 & 26.3 & 62.8 \\
\textbf{Top-P Threshold} & 83.3 & 85.3 & 86.5 & 83.6 & 26.8 & 62.9 \\
\bottomrule
\end{tabular}
}
\caption{Component ablation study. Accuracy Retention = (Variant Score / Dense Baseline) $\times$ 100. Latency measures total time for generating 1024 tokens. Top-P=0.5 serves as static threshold baseline.}
\label{table:component}
\end{table*}

\begin{table*}[ht]
\centering
\scalebox{0.93}{
\begin{tabular}{lcccc>{\columncolor{green!5}}cc}
\toprule
\textbf{Models} & \textbf{Prompt Len} & \textbf{Generation Len} & \textbf{Dense} & \textbf{Griffin} & \textbf{CLADA} & \textbf{Diff to Griffin} \\
\midrule
LLaMA-3-8B & 1024 & 1024 & 80.16 & 63.32 & 65.13 & -1.68 \\
(\textit{context\_len\_limit=8K}) & 2048 & 1024 & 84.34 & 68.08 & 68.09 & -0.01 \\
 & 2048 & 2048 & 87.92 & 75.86 & 76.37 & -0.51 \\
 \rowcolor{green!5}~& 4K & 2048 & 92.08 & 78.03 & 77.78 & 0.25 \\
 & 8K & 2048 & 96.53 & 84.00 & 83.79 & 0.21 \\
\midrule
Mistral-7B & 1024 & 1024 & 79.28 & 63.26 & 64.94 & -1.68 \\
(\textit{context\_len\_limit=32K}) & 2048 & 1024 & 81.37 & 67.48 & 68.03 & -0.55 \\
 & 2048 & 2048 & 85.46 & 75.74 & 76.15 & -0.41 \\
 \rowcolor{green!5}~& 4K & 2048 & 89.26 & 77.46 & 77.41 & 0.05 \\
 & 8K & 2048 & 93.43 & 83.79 & 83.22 & 0.57 \\
 & 16K & 2048 & 97.68 & 86.41 & 84.50 & 1.91 \\
 & 32K & 2048 & 102.35 & 91.35 & 89.78 & 1.57 \\
\bottomrule
\end{tabular}
}
\caption{Long-context efficiency comparison(latency in seconds). CLADA outperforms Griffin by up to $1.9s$ at maximum context lengths.}
\label{table:4k}
\end{table*}

\subsection{Ablation Studies}
We conduct systematic ablation studies to quantify the contributions of CLADA's core components. All experiments maintain identical hyperparameters and evaluation metrics as the main experiments. Ablation study on larger batch size can be seen in Appendix~\ref{app: batch size}.

\paragraph{Component Contribution Analysis}
Table~\ref{table:component} reveals three key findings through controlled component removal:

\begin{itemize}
\item \textbf{Statistical Sparsity}: Disabling prefix-based activation patterns (\textbf{w/o Stat. Sparsity}) causes $17.6-23.7\%$ performance degradation across tasks, while only reducing latency by $1.8-3.2\%$. This demonstrates its critical role in preserving model capacity through sequence-level regularities.
    
\item \textbf{Semantic Adaptability}: Removing cognitive-load-aware adjustments (\textbf{w/o Sem. Adapt.}) maintains latency parity but reduces accuracy by $2.4-3.4\%$, confirming its necessity for handling high-entropy contexts.

\item \textbf{Dynamic Thresholding}: Replacing our adaptive thresholds with static Top-P=0.5 (\textbf{Top-P Threshold}) leads to 14.7-$16.3\%$ performance drops, validating the superiority of our theoretically grounded approach.
\end{itemize}

\paragraph{Context Length Scalability}
Table~\ref{table:4k} demonstrates CLADA's superior scalability to long contexts compared to Griffin. While both methods exhibit latency growth with increasing context length, CLADA achieves $1.57-1.91s$ speedup at 32K context due to better memory access patterns in our dynamic thresholding strategy.



\section{Conclusion \& Discussion}
Our work introduces \textbf{C}ognitively-\textbf{L}oad \textbf{A}ware \textbf{D}ynamic Activation(\textbf{CLADA}), a novel framework that unifies statistical sparsity and semantic adaptability for efficient LLM inference. The key findings are as follows: 
\paragraph{Theoretical Insights} We identify and formalize two complementary properties of LLMs’ activation patterns:
\begin{enumerate}
    \item \textbf{Statistical Sparsity}: Activation patterns are dominated by prefix, providing a global efficiency baseline.
    \item \textbf{Semantic Adaptability}: High-cognitive-load contexts(e.g., unanticipated terms, complex syntax) trigger dynamic adjustments in activation, preserving local precision.
\end{enumerate}

\paragraph{Inference Gains} CLADA combines offline-searched TT thresholds with cognitive-load-aware adjustments, achieving 18-25\% speedup with less than 2\% performance degradation across six LLMs and nine tasks.

For \textbf{Future Directions} and \textbf{Broader Impact}, please see Appendix~\ref{sec: more conclusion}.


\section*{Limitations}
Despite its advantages, CLADA has several limitations that warrant further investigation:
\paragraph{Pre-filling Overhead} For long prefixes (>2048 tokens), the pre-filling phase accounts for 10-15\% of the total inference time, reducing the net speedup in real-time applications.  
\paragraph{Hardware Constraints} Storing activation masks for large models (e.g., LLaMA-70B) increases GPU memory usage, limiting deployment on resource-constrained devices.

\bibliography{custom}

\appendix

\section{Related Work}\label{app: related work}
\subsection{The Hint of Sparsity}
LLMs often appears excessive activation of neurons during tasks, leading to inefficiency and wasted resources\cite{bommasani2022opportunitiesrisksfoundationmodels, yuan2024llminferenceunveiledsurvey}. Studies\cite{liu2023modelbasedcontrolsparseneural} show that dense neural networks often display surplus activation. Treating sparsity as a continuous process can optimize model architecture holistically. The Lottery Hypothesis\cite{frankle2019lotterytickethypothesisfinding, malach2020provinglotterytickethypothesis} highlights pruning techniques to remove unnecessary connections and leverage inherent sparsity.

\subsection{Static Sparsity}
Early efforts to improve LLM efficiency focused on \textbf{static sparsity}, where redundant parameters are permanently removed or compressed.

\paragraph{Weight Pruning} Methods like SparseGPT\cite{frantar2023sparsegptmassivelanguagemodels} prune weights based on magnitude criteria, achieving up to 50\% sparsity without retraining. However, such approaches ignore input-specific sparsity patterns, leading to suboptimal performance on dynamic tasks like dialogue generation. Static methods lack adaptability to varying input contexts and often require retraining to recover performance.  

\subsection{Dynamic Activation}
Dynamic activation(DA) methods selectively activate subsets of neurons during inference, offering a balance between efficiency and flexibility. Existing researches on dynamic activation methods can be categorized in Table \ref{table: two types of da}.

\paragraph{Training-Dependent Methods}
DejaVu\cite{liu2023dejavucontextualsparsity}predicts activation sparsity during training by leveraging ReLU’s inherent sparsity. While achieving 20\% speedup on OPT-style models, it fails on non-ReLU architectures like LLaMA’s SwiGLU.

MoEfication\cite{zhang2022moeficationtransformerfeedforwardlayers} dynamically routes inputs to expert subnets, but requires costly expert training and introduces routing overhead. DS-MoE\cite{pan2024densetrainingsparseinference} introduces a framework that employs dense computation during training and switches to sparse computation during inference. LLaMA-MoE\cite{zhu2024llamamoebuildingmixtureofexpertsllama} offers a new lightweight method to transform FFNs into MoEs. LTE\cite{zheng2024learnefficientbuildstructured} achieves a superior balance between sparsity and performance by activating fewer neurons and is applicable to models with both ReLU and non-ReLU activation functions.
Lory\cite{zhong2024loryfullydifferentiablemixtureofexperts} retains the autoregressive properties of language models by adopting a causally segmented routing strategy and a similarity-based data batching method. This enables efficient expert merging operations and promotes specialization among experts in processing similar documents during training sessions.

\begin{table*}[h]
\begin{threeparttable}
    \centering
    \scalebox{0.7}{
    \begin{tabular}{c|p{7cm}p{3cm}p{2cm}p{4cm}}
    \toprule
    \textbf{DA Types} & \textbf{Definetions} & \textbf{Examples} & \textbf{Advantages} & \textbf{Current Limitations} \\
    \midrule
    \multirow{2}{*}{Training-Dependent DA} & Some leverage a \textit{predictor}, which is pre-trained using the model's training data, to dynamically identify essential activation neurons or experts during the model's forward. (Figure \ref{figure:Training-Dependent DA}) & DejaVu \cite{liu2023dejavucontextualsparsity}, MoEfication\cite{zhang2022moeficationtransformerfeedforwardlayers} & High Sparsity & Tend to underperform on models with non-ReLU activations(See Table~\ref{table: dejavu}) \\
    \cmidrule(lr){2-5}
    ~ & Others aim to reduce computational costs by employing multi-stage MoE-style training and introducing efficiency and separability loss penalties. & LTE \cite{zheng2024learnefficientbuildstructured} and D2DMoE \cite{szatkowski2024exploitingactivationsparsitydense} & High performance & Extra training required \\
    \midrule
    Training-Free DA & Employs pre-searched or pre-defined thresholds or sparsity levels to decide which neurons to retain or discard. Neurons with activation values falling below this bar are eliminated during current forward, thereby reducing computational overhead and latency.(Figure \ref{figure:Training-Free TDA}) & Griffin\cite{dong2024promptpromptedadaptivestructuredpruning}, CLADA(Ours) & Training-free for all model archs & Lower performance \\
    \bottomrule
    \end{tabular}
    }
    \caption{Two types of DA methods}
    \label{table: two types of da}
\end{threeparttable}
\end{table*}

\begin{figure*}[htbp]
    \begin{minipage}[t]{0.5\linewidth}
        \centering
        \includegraphics[height=2.7in]{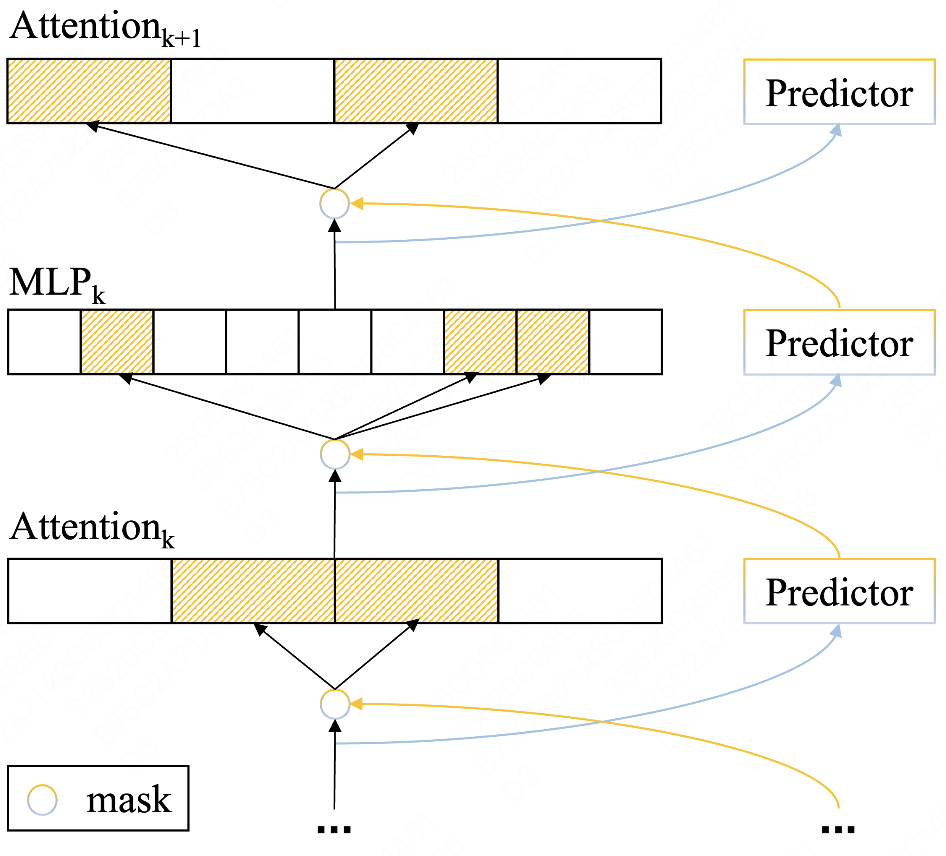}
        \caption{Training-Dependent DA}
        \label{figure:Training-Dependent DA}
    \end{minipage}
    \begin{minipage}[t]{0.5\linewidth}
        \centering
        \includegraphics[height=2.7in]{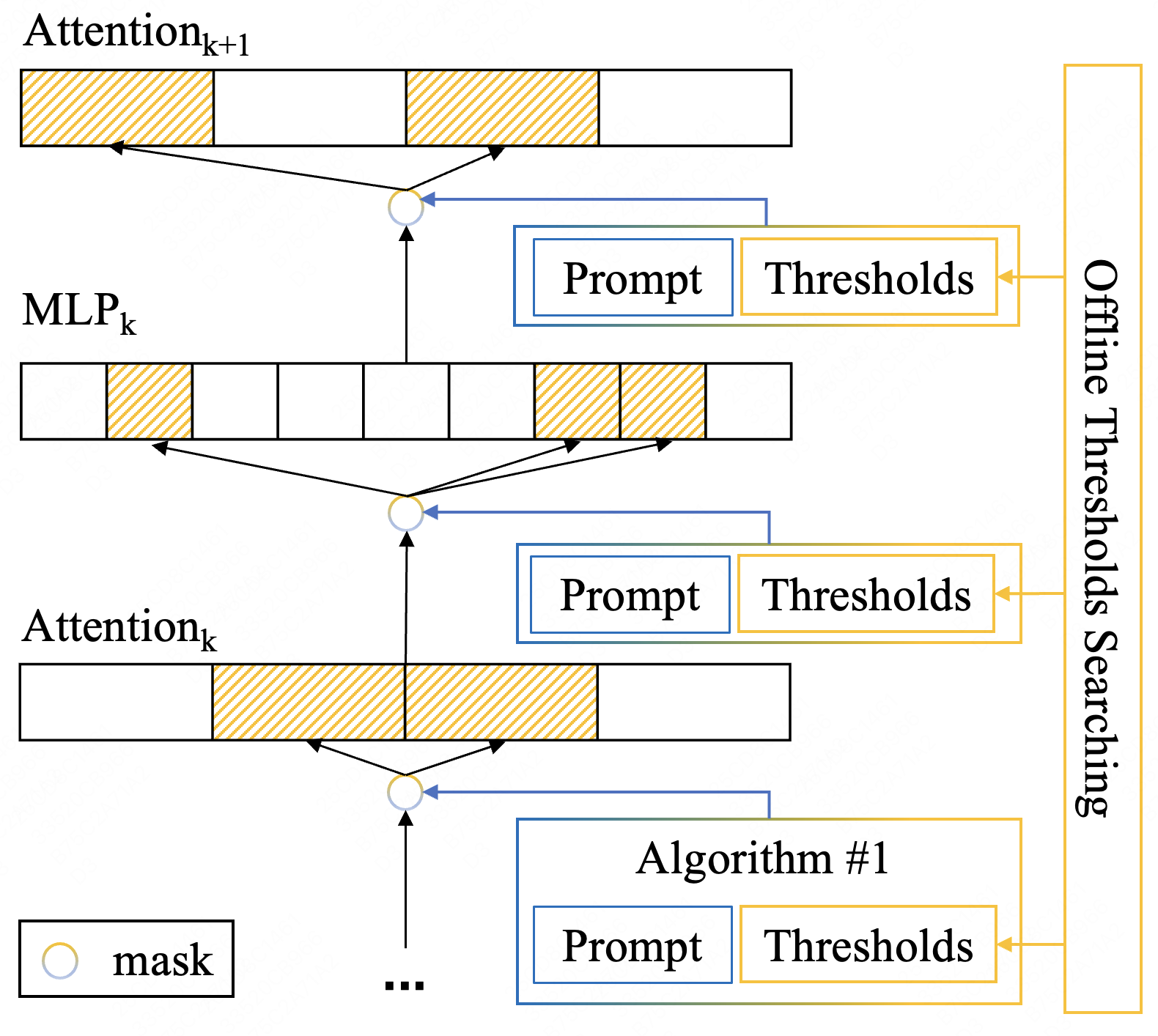}
        \caption{Training-Free DA}
        \label{figure:Training-Free TDA}
    \end{minipage}%
\end{figure*}
     
\paragraph{Training-Free Methods}
Griffin\cite{dong2024promptpromptedadaptivestructuredpruning} uses sequence-level activation clustering (flocking) to skip redundant computations only in generation phase. Despite its simplicity, Griffin suffers from significant performance drops (>3\% on QA tasks) due to heuristic threshold selection.

\subsection{Cognitive Load and Language Modeling}
Cognitive studies on human language processing provide inspiration for efficient computational models. Understanding how humans process language can lead to the development of more efficient and accurate language models. In this section, we explore the relationship between cognitive load, as measured by ERP components such as N400 and P600, and language modeling metrics like surprisal and entropy.

\paragraph{Surprisal and Entropy}
Surprisal, proposed by \citet{hale-2001-probabilistic}, is defined as the negative log probability of a word given its context. Surprisal captures the idea that the cognitive effort required to process a word is proportional to its unpredictability in a given context. This theory has been widely used to predict human reading times and has been shown to correlate with N400 amplitudes\cite{salicchi-hsu-2025-every}.

Entropy\cite{oh2024frequencyexplainsinversecorrelation}, on the other hand, measures the level of uncertainty about the upcoming linguistic input at a given point in a sentence. It is calculated based on the probability distribution of possible words in a context and has been linked to the complexity of language processing.

\section{Details on Bridging Cognitive Load and Activation}\label{sec: details}

\begin{table}[htbp]
\centering
\scalebox{0.8}{
\begin{tabular}{r|cccc}
\toprule
~ & Mean & Std & Min & Max \\
\midrule
Prefix\_len & - & - & 512 & 1024 \\
Surprisal & 0.58 & 0.31 & 0.01 & 1.01 \\
Entropy & 0.42 & 0.45 & 0.01 & 1.01 \\
Token\_len & 2048 & 0 & 2048 & 2048 \\
\bottomrule
\end{tabular}
}
\caption{Descriptive statistics of linear regression variables. The \textit{Surprisal} and \textit{Entropy} values in the table are sequence-level normalized to eliminate the impact of inter-sequence variability and absolute magnitude differences.}
\label{table: descriptive}
\end{table}

\begin{table}[ht]
\centering
\scalebox{0.8}{
\begin{tabular}{lccccc}
\toprule
\textbf{Batch\_size} & \textbf{1} & \textbf{2} & \textbf{4} & \textbf{8} & \textbf{16} \\
\midrule
\textbf{Latency} & 64.03 & 74.00 & 76.36 & 79.33 & 82.85 \\
\bottomrule
\end{tabular}
}
\caption{Batch size scalability analysis. Latency measures total processing time.}
\label{tab:batch}
\end{table}

\subsection{Why Token-2049?}\label{app: token}
Selecting token-2049 is imperative for the two reasons: 1) The CLADA method introduced in this paper facilitates inference speedup. Thus, it is crucial to examine the activation of the $2049_{th}$ or latter token when the prefix length is 2048. 2) While analyzing the $2049_{th}$ token, as opposed to the $(prefix+1)_{th}$ token, may diminish the precision of observed surprisal and entropy effects, it more accurately adheres to the hypothesis of activation flocking.

\subsection{Visual Evidence}\label{app: visual}
\begin{figure*}[h]
    \begin{minipage}[t]{0.33\linewidth}
        \centering
        \includegraphics[height=1.35in]{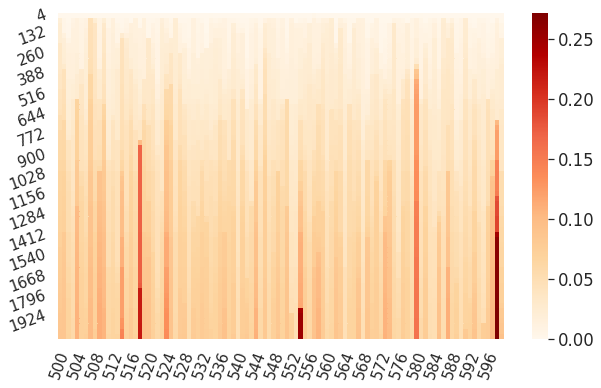}
        \caption{Activation of NLS-A}
        \label{figure:nlsa}
    \end{minipage}
    \begin{minipage}[t]{0.33\linewidth}
        \centering
        \includegraphics[height=1.35in]{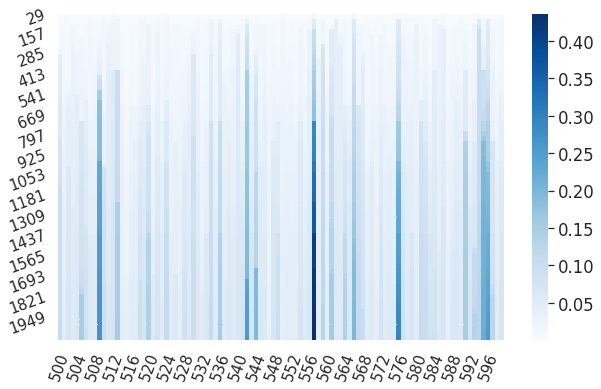}
        \caption{Activation of NLS-B}
        \label{figure:nlsb}
    \end{minipage}
        \begin{minipage}[t]{0.33\linewidth}
        \centering
        \includegraphics[height=1.35in]{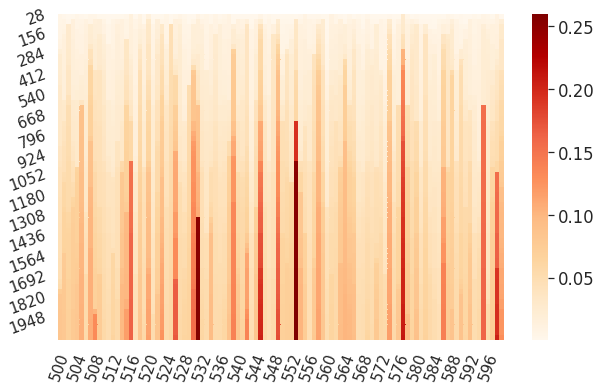}
        \caption{Activation of NLS-A'}
        \label{figure:nlsa'}
    \end{minipage}
\end{figure*}

We plot the whole sequence activation heat-map of NLS-A/B/A' separately in Figure~\ref{figure:nlsa}, Figure~\ref{figure:nlsb}, and Figure~\ref{figure:nlsa'}.

In these figures, the horizontal axis represents the neuron indices, while the vertical axis represents the token indices.

Figures \ref{figure:nlsa} to \ref{figure:nlsa'} demonstrate that substituting the initial 512 tokens of NLS-A with the prefix from NLS-B results in an activation pattern that more closely aligns with that of NLS-B.

\subsection{Case Study}\label{app: detailed samples}
The mechanism behind statistical sparsity phenomenon needs detailed investigation. When processed as a sequence, activated neurons consider all tokens, suggesting that statistical sparsity may be due to preceding tokens rather than the current token.

To eliminate sequence information influence, we selected specific samples and conducted a similarity analysis of their activation patterns. Details in Table~\ref{table: detailed 13 samples} in Appendix~\ref{app: detailed samples}.

\begin{table*}[p]
\centering
\scalebox{0.8}{
\begin{tabular}{c|ll}
\toprule
\textbf{Index} & \multicolumn{1}{c}{\textbf{Samples}}                               & \multicolumn{1}{c}{\textbf{Treatments}}          \\
\midrule
1              & "\#\#\# Article: Almost one million people visited the city"       & Baseline                                   \\
2              & "Article: Almost one million people visited the city"              & Remove beginning token                     \\
3              & "Almost one million people visited the city"                       & Remove beginning tokens                    \\
4  & "\#\#\# Article: Nearly one million people visited the city" & Modify the word at the beginning of the sequence. \\
5              & "Nearly one million people visited the city"                       & Remove beginning tokens                    \\
6              & "\#\#\# Article: Less than one million people visited the city"    & Change to antonym                          \\
7              & "Less than one million people visited the city"                    & Remove beginning tokens                    \\
8              & "\#\#\# Article: Almost one million people visited the city"       & Similarity threshold                          \\
9              & "\#\#\# Article: Almost one million people visited the restaurant" & Change to synonyms                        \\
10 & "Almost one million people visited the restaurant"           & Modify the word at the end of the sequence        \\
11             & "Almost one million people visited the planet"                     & Modify the word at the end of the sequence \\
12 & "Almost one million tourists visited the restaurant"         & Modify the words at the middle and end            \\
13             & "Almost one million aliens visited the planet"                     & Dissimilarity threshold    \\        
\bottomrule
\end{tabular}
}
\caption{Detailed 13 samples for activation statistical sparsity check.}
\label{table: detailed 13 samples}
\end{table*}

Table~\ref{table: detailed 13 samples} details the 13 samples used for activation pattern similarity analysis. Samples 1-3 and Samples 4, 6, and 9 form two treatment groups. If Sample 4 shows greater similarity to Sample 1 than to Samples 2 and 3, it supports Claim~\ref{final claim}.

\begin{figure*}[p]
\centering
\includegraphics[scale=0.75]{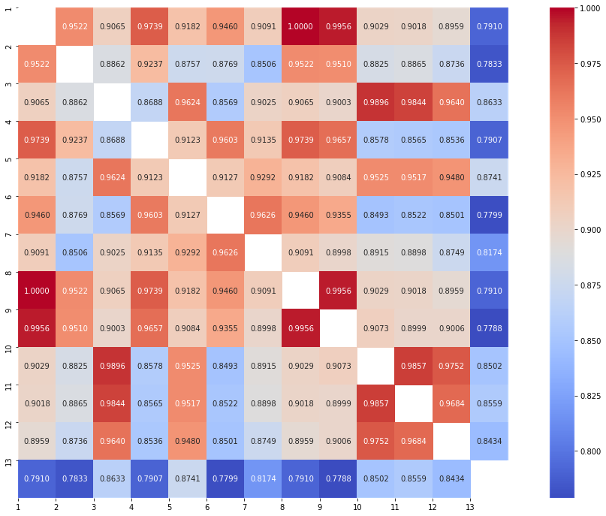}
\caption{Similarity matrix of 13 samples' activation pattern}
\label{figure: similarity}
\end{figure*}

From the similarity heatmap in Figure~\ref{figure: similarity}, we observe the following:
a) Samples 4, 6, and 9 are more similarly activated to Sample 1 than to Samples 2 and 3;
b) Samples 1, 6, 8, and 9 are more similarly activated to Sample 4 than to Sample 5;
c) Sample 9 is more similarly activated to Samples 4, 6, and 8;
d) Samples 11 and 12 are more similarly activated to Samples 9 and 10;
e) Samples 10, 11, and 12 are more similarly activated to Sample 13 than to any other samples.

\section{Batch Size Analysis}\label{app: batch size}
We evaluate CLADA's scalability across different batch sizes using LLaMA-3-8B on XSum summarization. The experiments are conducted on a single NVIDIA A100 GPU with 80GB memory, using FP16 precision. We measure total time for processing the batch. Table~\ref{tab:batch} indicated that processing time increases sublinearly($27.6\%$ from batch 1 to 16) due to optimized memory access patterns in our dynamic activation strategy.

\section{More Conclusion}\label{sec: more conclusion}
\subsection{Future Directions}
To address these limitations and extend the impact of CLADA, we propose the following future directions: 
\begin{enumerate}
    \item Cross-Modal Sparsity: Extend CLADA to multimodal models (e.g., LLaVA) by incorporating visual surprisal and entropy metrics for joint text-image processing.
    \item Hardware Optimization: Design lightweight mask compression algorithms (e.g., sparse encoding) to reduce memory footprint and enable edge deployment.
\end{enumerate}

\subsection{Broader Impact}
Beyond efficiency improvements, CLADA offers a new perspective on bridging cognitive science and computational models. By formalizing the connection between cognitive mechanisms and LLM activation patterns, our work paves the way for more interpretable and biologically plausible AI systems. Furthermore, the training-free nature of CLADA makes it accessible for a wide range of applications, from real-time dialogue systems to on-device language processing. 

\end{document}